\documentclass[conference]{IEEEtran}
\IEEEoverridecommandlockouts
\usepackage{cite}
\usepackage{amsfonts}
\usepackage{textcomp}
\usepackage{xcolor}
\def\BibTeX{{\rm B\kern-.05em{\sc i\kern-.025em b}\kern-.08em
    T\kern-.1667em\lower.7ex\hbox{E}\kern-.125emX}}

\usepackage[final]{graphicx}
\graphicspath{ {./} }
\usepackage[reqno]{amsmath}
\usepackage{amssymb}
\usepackage{amsthm}
\usepackage{subfig}
\usepackage{epstopdf}
\usepackage{algorithm}
\usepackage{algorithmicx}
\usepackage{algpseudocode}
\usepackage{setspace}
\usepackage[T1]{fontenc}
\usepackage{color,soul}
\usepackage{multirow}
\usepackage{array}
\usepackage{cite}
\usepackage{verbatim}
\usepackage{enumitem}
\usepackage{url}
\usepackage{csvsimple}
\usepackage{float}

\usepackage{float}
\usepackage{graphicx}
\usepackage{caption}
\usepackage[capitalize]{cleveref}
\usepackage{environ}

\usepackage{array}
\newcommand{\PreserveBackslash}[1]{\let\temp=\\#1\let\\=\temp}
\newcolumntype{C}[1]{>{\PreserveBackslash\centering}p{#1}}
\newcolumntype{R}[1]{>{\PreserveBackslash\raggedleft}p{#1}}
\newcolumntype{L}[1]{>{\PreserveBackslash\raggedright}p{#1}}

\begin{document}
\raggedbottom

\title{Data-Driven Subsampling in the Presence of an Adversarial Actor}
 
\author{\IEEEauthorblockN{Abu Shafin Mohammad Mahdee Jameel\IEEEauthorrefmark{1}, Ahmed P. Mohamed\IEEEauthorrefmark{1}, Jinho Yi\IEEEauthorrefmark{1}, Aly El Gamal\IEEEauthorrefmark{1} and Akshay Malhotra\IEEEauthorrefmark{2} }
\IEEEauthorblockA{\IEEEauthorrefmark{1} School of Electrical and Computer Engineering, Purdue University, USA}
\IEEEauthorblockA{\IEEEauthorrefmark{2}InterDigital Communications, Inc., USA}
\IEEEauthorblockA{Email: {\{amahdeej, mohame23, yi62, elgamala\}@purdue.edu, akshay.malhotra@interdigital.com}}
\thanks{The first two authors contributed equally to this paper.}
}

\maketitle

\begin{abstract}

Deep learning based automatic modulation classification (AMC) has received significant attention owing to its potential applications in both military and civilian use cases. Recently, data-driven subsampling techniques have been utilized to overcome the challenges associated with computational complexity and training time for AMC. Beyond these direct advantages of data-driven subsampling, these methods also have regularizing properties that may improve the adversarial robustness of the modulation classifier. In this paper, we investigate the effects of an adversarial attack on an AMC system that employs deep learning models both for AMC and for subsampling. Our analysis shows that subsampling itself is an effective deterrent to adversarial attacks. We also uncover the most efficient subsampling strategy when an adversarial attack on both the classifier and the subsampler is anticipated.\\


\end{abstract}
\IEEEpeerreviewmaketitle

\section{Introduction}

The use of adaptive modulation schemes, where multiple modulation types are utilized based on channel conditions or data types, have been shown to be more robust to noise and fading in wireless channels \cite{svensson2007introduction}. These schemes enhance the wireless system performance by reducing bit error rates (BER) and increasing the utilization of available link spectrum \cite{wang2020adaptive}. Adaptive modulation schemes have been incorporated in major current wireless protocols like IEEE 802.11n and 3GPP 5G NR \cite{wang2020adaptive}. For effective operation, adaptive modulation systems need an intelligent modulation classification scheme at the receiver that can detect the modulation type in real-time. Thus, automatic modulation classification (AMC) methods have been a topic of high interest. Different techniques based on average likelihood, power spectral density, frequency domain features, wavelet transform etc. \cite{dobre2007survey} have been utilized to design AMC systems. However, likelihood based approaches suffer from high computational complexity, and feature based approaches require manual design with expert domain knowledge \cite{meng2018automatic}. 

\begin{figure*}[ht]
    \centering
    \includegraphics[scale=0.27]{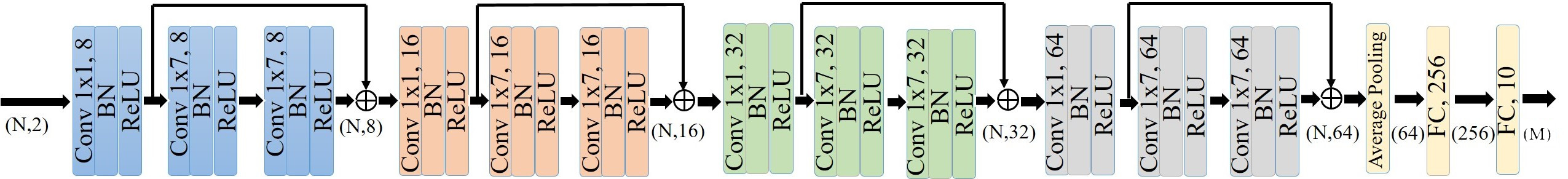}
    \caption{Architecture of the classifier Network.}
    \label{fig:ResNet_Architecture}
\end{figure*}

Due to their efficient computational capabilities, robustness, and good performance, recent research has focused on AMC using deep learning (DL) based systems \cite{ meng2018automatic, o2018over, zhang2023lightweight}. 
To further improve the performance of deep learning based AMCs and tackle the challenges of computation complexity, power, and training times, data-driven subsampling schemes have been utilized to pre-process the data before classification \cite{ramjee2021ensemble, ramjee2019fast}.The data-driven subsampling strategies utilize deep learning models that are trained to identify the optimal subsampling strategy to select a sub-set of data points from an input data vector \cite{shen2023multi}. 

One major concern regarding wireless communication in machine learning (ML) based wireless communication systems is the presence of an adversarial element. ML models that rely on stochastic gradient descent optimization are known to lack robustness against malicious adversarial examples that introduce small perturbations, specially crafted to cause ML models to malfunction when added to legitimate inputs \cite{goodfellow2014explaining}. In a wireless setting, a small perturbation would correspond to low power transmission performed by a jammer to introduce slight additional noise in the signal, that is difficult to detect at the receiver and leads to significant ML performance degradation. Recent studies have confirmed the vulnerability of ML models for modulation classification to such adversarial examples \cite{zolotukhin2022assessing}. This kind of data poisoning attack can introduce significant performance degradation in ML based classifiers \cite{usama2019adversarial}. More sophisticated attacks have also been demonstrated that take into account the noise encountered during actual over-the-air transmission of the perturbed signals \cite{kim2020over}. There has also been work towards designing robust AMC systems that can perform well even in the presence of adversarial attacks \cite{zhang2022hybrid, lee2023uniqgan}. 

In this paper, we investigate the effect of adversarial attacks on a data-driven subsampling based automatic modulation classification system. The combined system has two deep learning models, one performs the data-driven subsampling, while the other is responsible for the classification task. Having a setup with two deep learning models presents interesting possibilities for different levels of attacker knowledge. The contributions of this paper are as follows:

\begin{itemize}
    \item We demonstrate that introduction of data-driven subsampling makes a DL based AMC system more robust to adversarial attacks. \textbf{To the best of our knowledge, this is the first demonstration of adversarial robustness of data-driven subsampling.} This is true not only in the case when the attacker has complete knowledge about only the classifier DL model, but also in the case when the attacker has complete knowledge about both the classifier and subsampler DL model.
    \item We evaluate the robustness of subsampling based AMC strategies in the presence of an adversarial attack. This includes both fixed and data-driven subsampling strategies, and an ensemble method combining both approaches. \textbf{We further identify the subsampling strategy that provides the most robustness against adversarial actors.}
\end{itemize}

This paper is organized as follows. First, we describe in detail the structure of the deep learning based modulation classifier, and the data-driven subsampling algorithms considered for this work. Then, we explain the setup of the communication environment with the presence of an adversarial actor, and discuss the different level of capabilities an attacker might have. Finally, we present and analyze the findings of our investigations.

\begin{figure}[t]
    \centering
    \includegraphics[width=0.45\textwidth]{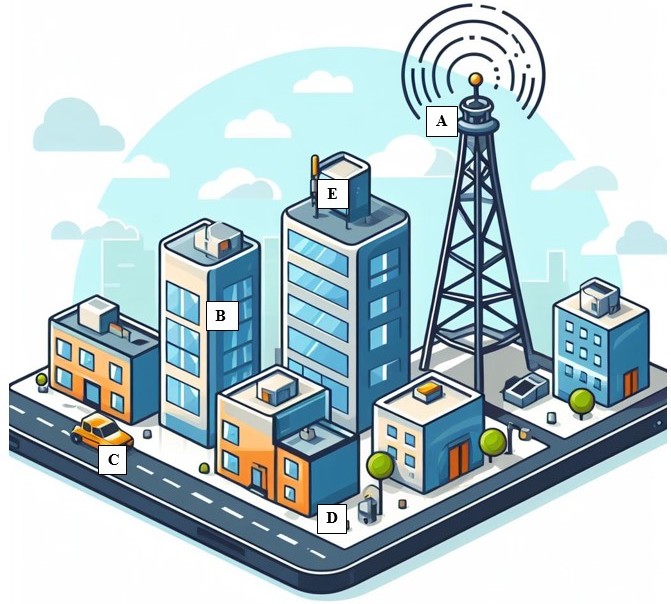}
   
    \caption{A typical communication scenario with an adversarial actor. (A) represents the base station. (B), (C), and (D) are indoor, outdoor, and vehicular User Equipment (UE), respectively. (E) is an adversarial base station that has the capability to intercept, modify and re-transmit data coming from (A). }
    \label{fig:system}
\end{figure}

\begin{figure*}[t]
    \centering
    \includegraphics[width=1\textwidth]{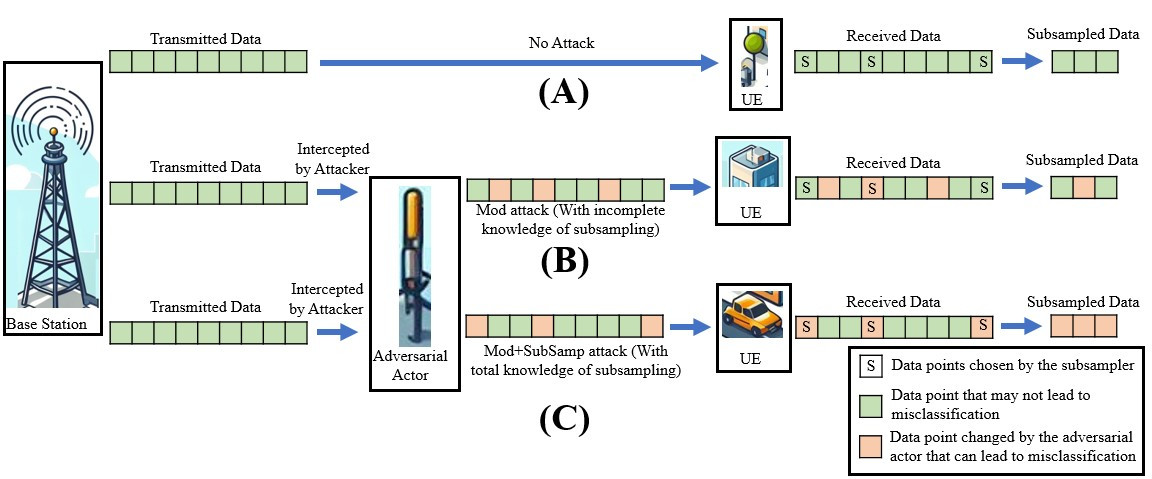}
   
    \caption{Possible communication scenarios from the base station to a UE in the presence of an adversarial actor: (A) No Attack, (B) Mod Attack, (C) Mod+SubSamp Attack.}
    \label{fig:threats}
\end{figure*}

\section{System Architecture} \label{section:sysarch}
In this section, we describe the automatic modulation classifier employed in our simulations. We also describe in detail the subsampling schemes used to determine the optimal subset of data points from an input sample.

\subsection{Deep Modulation Classifier}  \label{subsection:classifier}

We consider a wireless communication system consisting of a transmitter and a receiver. The transmitter encodes the data with one of $M$ available modulation schemes and transmits in-phase and quadrature (I/Q) domain samples over-the-air, while the receiver utilizes a deep learning model to predict the classification method based on the received I/Q samples and decode the data. 

For the classifier, we utilize a modified ResNet architecture represented in Fig. \ref{fig:ResNet_Architecture}. This architecture is similar to the one utilized in \cite{courtat2021light}, adding batch normalization layer after each convolutional layer to prevent overfitting. It accepts an input of size $(1, 2, N)$ representing (channel, I/Q samples, number of samples). For an input sample of length $L$, $N=L$ if no subsampling algorithm is present. When subsampling is employed $N=L/\alpha$, where $\alpha$ is an integer whose value changes based on the subsampling rate. The network utilizes four different residual blocks, each with different filter outputs, followed by an average pooling layer and two fully connected layers. Each residual block consists of three convolution layers. Inside each residual block, the output of the first convolutional layer is added to the output of the third convolutional layer using a skip connection. The final One-Hot output has $M$ components, each representing one of the possible $M$ modulation schemes. 

\subsection{Subsampling Schemes}\label{subsection:subsamplers}
The target of a subsampling scheme is to reduce the number of data points in an input intelligently, so that the reduction in input size does not negatively impact the classifier performance. Consider an I/Q sample with a length $L$. The target of the subsampling scheme is to reduce the length to $\{N: N = L/\alpha, \alpha \in factors(L)\}$. There are two different approaches to subsampling, fixed and data-driven.

\subsubsection{Fixed subsampling scheme} 

In this scheme, the $N$ data points are chosen according to some predetermined formula, and does not use any data for training.

\subsubsection{Data-driven subsampling scheme} \label{subsection:data-driven-subsamplers}

In this scheme, training data is used to determine a subsampling method that is best suited to this particular dataset. This is done by using a specially designed model called the Ranker Model (RM). The RM is a DL model that is pre-trained to perform modulation classification. 

Initially, we train the RM using I/Q samples of length $L$. We then iteratively simulate the removal of a specific data point in an I/Q sample by setting both the real and imaginary parts of that data point to zero. This modified data is input into the RM, effectively deactivating specific input neurons, and the weight from these neurons do not contribute to the outcome of the model. We evaluate the model's performance for all $L$ data points, resulting in $L$ classification accuracies from the ranker model. The most important data point is the one whose removal results in the lowest classification accuracy. 
   
Then, we set this most important data points to zero for the whole training set and train the RM again, with the target of finding the most important data point among the remaining $L-1$ data points. Then we repeat the process and continue until we get a set of $N$ highest ranked data points. 

In this work, we employ and compare both fixed and data-driven subsampler schemes. We also employ an ensemble method that utilizes multiple data-driven subsampling models to find the best subsample positions.

\setcounter{subsubsection}{0}
\subsubsection{Uniform Subsampler} This is a fixed subsampling scheme. This is a simple algorithm where the input data is sampled at uniform intervals. We get $N$ input data points from an I/Q sample of length $L$ by taking data points from position $\{d_k: k= \alpha, 2\alpha,  ....,  N\alpha\}$, where $\alpha=L/N$.

\subsubsection{Complex-CNN Subsampler} This is the first of the three data-driven subsampler schemes. The three schemes are separated by their choice of the ranker model. In this first case, the Complex-CNN network proposed in \cite{krzyston2020complex} is selected as the RM. Then we apply the data driven subsampling algorithm outlined in subsection \ref{subsection:data-driven-subsamplers} above. The Complex-CNN network has two specially designed convolution neural network (CNN) layers that are able to handle complex I/Q sample inputs. This architecture has been shown to significantly improve AMC performance. To our knowledge, this is the first time this network is employed as a RM for a subsampler.

\subsubsection{ResNet Subsampler} For this data-driven subsampler scheme, we use the same network as the classifier network described in section \ref{subsection:classifier} as the ranker model. This model has been shown to achieve superior performance while offering lower computational complexity.

\subsubsection{CLDNN Subsampler} This data-driven subsampler scheme uses a ranker model comprising of CNN and long short-term memory (LSTM) layers, as proposed in \cite{ramjee2021ensemble}. This model utilizes the correlations present in received I/Q data samples along the time axis.

\subsubsection{Holistic Subsampler} This is an ensemble method utilizing the three data-driven subsampling methods (Complex-CNN, ResNet and CLDNN Subsampler) presented above. After getting the list of $N$ highest ranked data points from the three subsamplers, we end up with a list of $\{P: P\leq3N \}$ data points. Then we choose the best $N$ data points from these $P$ points using an iterative data-driven approach, where we rank the data points by their selection by multiple subsamplers, and their impact on the RM classification accuracy.

\section{Communication Network Setup}

In Fig. \ref{fig:system}, we present a simplified example of a fictional city block representing the overall communication scenario. In this example, we have a mobile base station (A) transmitting to three User Equipment (UE) receivers. One is an indoor UE, one is outdoor and the last one is a vehicular UE. The base station employs an adaptive modulation scheme. The receivers employ the AMC scheme with subsampling, as outlined in Section \ref{section:sysarch}. We also have an adversarial actor (E) present in this fictional city block\textemdash who has the capability to intercept, modify and re-transmit the data sent from the base station to the UEs\textemdash with the intention of deceiving the DL based AMC present in a UE.

\subsection{Threat Models}

In the communication environment present in Fig. \ref{fig:system}, there are three possible communication scenarios. These scenarios are detailed in Fig. \ref{fig:threats}, where we use a simplified representation of the transmitted I/Q data sample to illustrate the different attack scenarios.

\subsubsection{No Attack} In this case the transmitted I/Q samples are not captured by the adversarial actor and reach the UE without any modification. The UE uses a subsampling scheme to select the best subset of the I/Q samples. 

\subsubsection{Mod Attack} The adversarial actor intercepts the I/Q frame sent by the base station. The adversary knows that the UE is employing a DL based AMC, and has complete knowledge about how the AMC works. It also has access to the same training data used to train the AMC for the UE. Using this information, the adversarial actor perturbs the I/Q frame in such a way that the AMC at the UE will misclassify it. Here the adversarial actor is missing a crucial piece of information about the subsampling scheme. It knows that a subsampling scheme is used, and it knows that $N$ data-points will be chosen from the I/Q sample of length $L$. It also has access to all five subsampling schemes outlined above, and can use the associated training data to train them. However, it does not know which subsampling scheme is employed by the UE, and thus is forced to perturb the data while assuming a randomly picked subsampling scheme.

\subsubsection{Mod+SubSamp Attack} In this scenario, the transmitted data is intercepted by the adversarial actor. The adversarial actor has complete knowledge of the deep learning classifier, including access to the training samples. It also has complete data about the subsampling scheme in operation. In the case of data-driven subsampling schemes, or the holistic subsampler, it has complete access to the ranker models and their associated training data. Using this data, the adversarial actor can predict the exact data points that will be picked up by the subsampler from the I/Q data frame. It makes changes to the data frame that are calculated to have the maximum chance of the UE subsampler and classifier model misclassifying the modulation scheme.

\begin{figure}[t]
    \centering
    \includegraphics[width=0.5\textwidth]{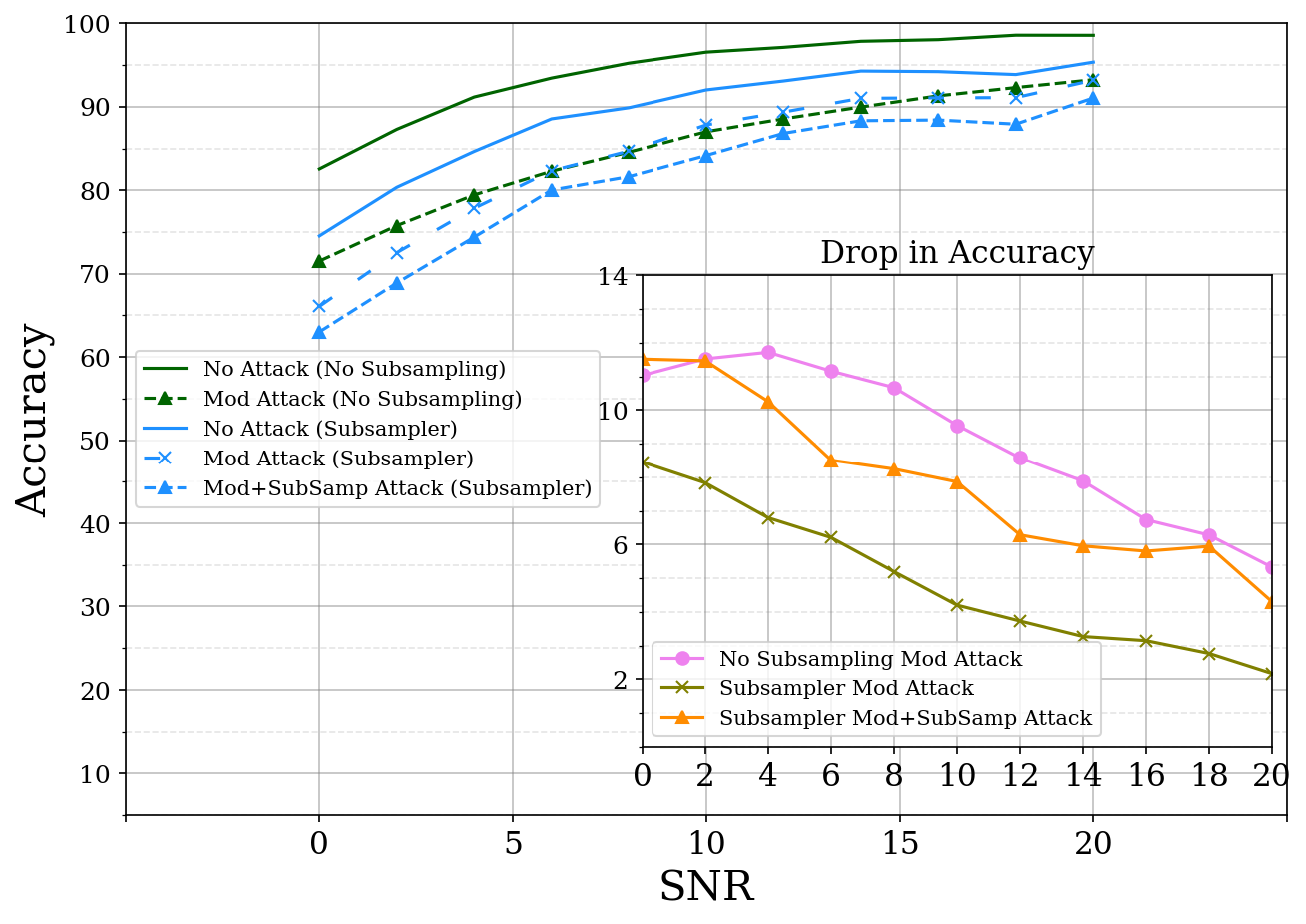}
    \caption{Comparison of classification accuracy for no subsampling vs ResNet subsampler ($\frac{1}{2}$ subsampling rate), in the presence of both Mod Attack and Mod+SubSamp Attacks. In inset, the accuracy difference between scenarios with no subsampling and subsampler is shown for different attack conditions.}
    \label{fig:robustness}
\end{figure}

\begin{figure*}[ht]
    \centering
    \includegraphics[width=1\textwidth]{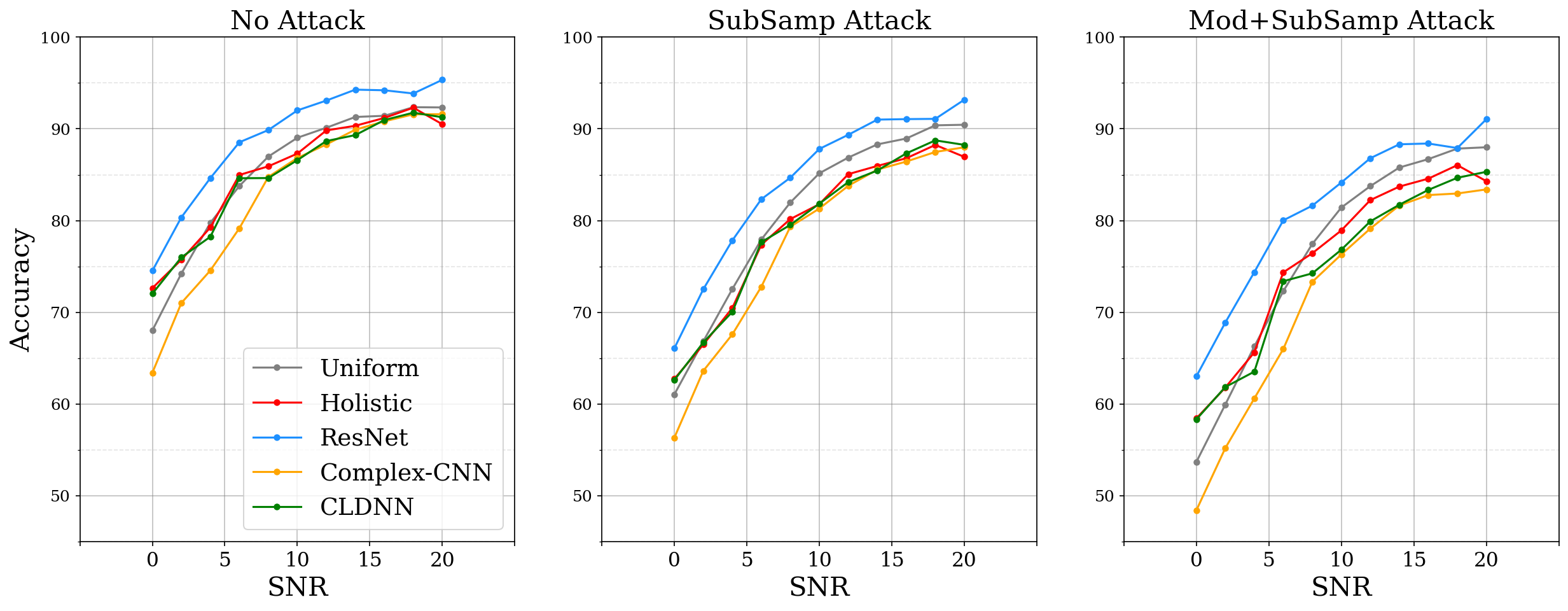}
    \caption{Performance of different subsamplers under Mod Attack and Mod+SubSamp Attack.}
    \label{fig:defense}
\end{figure*}

\subsection{Gradient-based Adversarial Attack}
In order to defeat a DL model (Both the modulation classifier, and the ranker model for the subsampler), the attacker employs the popular the popular Carlini-Wagner (CW) $L_\infty$ Attack \cite{carlini2017towards}, which has been shown to be effective against DL based methods. The adversarial actor generates $x$, an adversarial version of the intercepted data $x_{0}$ by solving the optimization problem that is formulated as:

\begin{equation}
    \min_x{\lVert x-x_0 \rVert}_\infty+c.f_t(x)
\end{equation}
where $f_t$ is defined by:

\begin{equation}
    f_t(x')=\max(\max (Z(x')_i: i\neq t) - Z(x')_t,0) 
\end{equation}
where ${\lVert x-x_0 \rVert}_\infty$ is $L_\infty$ distance that measures the maximum variation for any of coordinates as ${\lVert x-x_0 \rVert}_\infty = max(|x-x_{0_1},...,x-x_{0_N}|)$, $Z(.)$ is a softmax function, $Z(x')$ is the classifier model logit when the input considered is adversarial example $x'$, and $t$ is a target label.  $t$  is used for  minimizing (2), over all available labels excluding the true label. The perturbation norm is limited such that perturbation power is calculated as the noise power of the signal. 

This $x$, which is then re-transmitted and received by the UE, is able to fool the target DL model into misclassifying the modulation scheme.

\section{Dataset} 

We utilize a modulation classification dataset generated by the RML22 dataset generation code \cite{sathyanarayanan2023rml22}. This is an updated version of the popular RML16 dataset \cite{o2016radio}, providing a more realistic and carefully designed signal model parameterization. 

The modulated signals of the generated dataset are simulated in a Rayleigh fading channel environment with additive white Gaussian noise and variable delay spreads, Doppler shift, sample rate offset, center frequency offset and phase offset. The generated RML22 dataset consists of total 10 different modulation forms, 8 digital (BPSK, QPSK, 8PSK, PAM4, QAM16, QAM64, GFSK and CPFSK) and 2 analog (WBFM and AM-DSB) modulation forms, where each modulation form has 21 levels of uniformly distributed SNRs from -20 dB to 20 dB in 2 dB steps. The dataset consists of 1,260,000 sample examples. Each sample example is composed of 128 samples in length for two channels I/Q that can be represented as  2x128 samples. In this work, the dataset is distributed as 40\%, 10\% and 50\% for training, validation and testing sets, respectively.

\begin{table}[t]
        \centering
        \captionsetup{justification=centering}
	\caption{Impact of Attacks on Classification Accuracy.}
	\label{table:robustness}
    \tabcolsep=0.05cm
    \begin{tabular}{c|c|c|c} 
    \hline
     & No Atck.& Mod Atck.& Mod+SSamp Atck.\\ 
    \hline
    No Subsampling&    98.4\%&     91.3\%&       \\
    Drop in Accuracy& & \textbf{6.74\%}&\\ 
    \hline
    Subsampling&     94.2\%&     91.06\%&      88.4\%\\ 
    Drop in Accuracy&    &    \textbf{3.14\%}&     \textbf{5.8\%}\\
    \hline
    \end{tabular}
    
\end{table}

\section{Experimental Results} 

Firstly, in Table \Ref{table:robustness}, we investigate the impact of subsampling on an adversarial actor. We present the accuracy of the system with and without subsampling under 16dB SNR. Here we consider the ResNet subsampler with ($N=64, \alpha = 2$). We then consider the scenario when an adversarial actor is present. In the no subsampling case, the adversary can only perform a Mod Attack. With the data-driven subsampler, we consider both the Mod Attack and the Mod+SubSamp Attack.

\begin{itemize}
     \item When there is no attack, the introduction of subsampling causes a minor performance penalty. As the classifier has half the number of data points available in each frame of I/Q samples, the accuracy drops. On the other hand, only having to process half the number of data points speeds up the classifier considerably.
     \item In case of Mod Attack, the drop in accuracy under attack is less than half when subsampling is present. The subsampling accuracy is now almost same as the no subsampling accuracy. We note that even in the case of Mod attack, the adversarial actor has most of the information about the subsampler except for  the exact knowledge of the subsampler that is deployed.
     \item In case of Mod+SubSamp attack, the drop in accuracy is still lower than that of no subsampling. This shows that even in the case of absolute knowledge for the attacker, subsampling retains its advantage in adversarial robustness.
\end{itemize}

Then in Fig. \ref{fig:robustness}, we present a more comprehensive analysis for the same cases across an SNR range from 0dB to 20dB. The drop in accuracies across this SNR range is presented in the inset of the plot. From the inset, we can clearly see that the drops in accuracy when having no subsampling is consistently the highest. In case of Mod Attack, there is a very big difference between the no subsampling accuracy drop and the subsampler accuracy drop. The Mod+SubSamp attack accuracy drops also show a big gap with the no subsampling accuracy drop.

This is extremely pertinent, as it shows that \textbf{subsampling imparts an inherent adversarial robustness}. It is to be noted that subsampling can be compared to data compression algorithms, and data  compression algorithms have been shown to be successful against adversarial attacks in the image processing domain \cite{das2018shield}.

\begin{table}
    \centering
    \captionsetup{justification=centering}
	\caption{Percentage change in classification accuracy under different attack scenarios.}
	\label{table:accr_avg}
    \tabcolsep=0.2cm
    \begin{tabular}{c|c|c} 
    \hline
    & Mod Attack& Mod+SubSamp Attack\\ \hline
    CLDNN          &  4.84&   9.77\\ 
    Complex-CNN            &  4.86&  11.5\\ 
    ResNet         &   \textbf{3.31}&   \textbf{8.55}\\ 
    Holistic       &   5.37&  9.61\\ 
    Uniform        &  3.92& 11.57\\ \hline
    \end{tabular}
\end{table}

Next, we focus on the impact of the choice of subsampler in an adversarial scenario. In Fig. \ref{fig:defense}, we present the classification accuracy under No Attack as well as Mod and Mod+SubSamp attack scenarios for all five subsamplers. Here we consider ($N=64, \alpha = 2$), and an SNR range from 0dB to 20dB. From the figures we can observe that the ResNet subsampler provides the best classification accuracies in all three scenarios. The behavior of the five subsamplers follow a clear pattern across the three scenarios, and we can easily identify better performing subsamplers.

Finally, in Table \ref{table:accr_avg}, we calculate the proportional percentage change in classification accuracy across different attack scenarios for the five subsamplers under consideration. Here we average out classification accuracies across the SNR range from 0dB to 20dB, and also average across three choices of subsampling rates: ($N=64, \alpha = 2$), ($N=32, \alpha = 4$), and ($N=16, \alpha = 8$). We can see that the ResNet subsampler has the lowest accuracy drop under attack for both the Mod attack scenario, and the Mod+SubSamp attack scenario. We note that the Mod attack accuracy drops are significantly smaller than the Mod+SubSamp attack accuracy drops in all cases.

Based on this analysis, we present the following recommendations for best nullifying adversarial attempts in a scenario where the presence of an adversarial actor is suspected:
\begin{itemize}
    \item In addition to having the best overall performance across a wide range of SNR values, the ResNet subsampler is also the most robust against both Mod and Mod+SubSamp attacks. It is the best subsampler in an environment where an adversarial actor is trying to force the UE into misclassification.
    \item In the absence of only a small information about the  specific subsampler being utilized, the adversarial actor's ability to impact the classification accuracy becomes significantly limited. This presents an interesting opportunity for the base station and UE to prevent such attacks. \textbf{By protecting this single piece of information about the choice of the subsampler, the adversarial robustness of the overall system can be significantly increased}, even when the adversary has access to the training data, full classifier network, and all possible ranker models. The communication system can be designed in such a way that the choice of the subsampler is given top priority and communicated with enhanced security. This also makes the availability of multiple subsampling choices crucial. A system can be designed in such a way that the subsampler is randomly changed over time, and this change is conveyed over a secure channel. This would severely limit the adversarial actor's effectiveness and improve the AMC classifier performance in an adversarial setup.
\end{itemize}

\section{Conclusion} 

In this paper, we explore the benefit of data-driven subsampling in introducing adversarial robustness for automatic modulation classification. We show that in the presence of an adversarial actor, classification accuracy is significantly improved by securing only the information about the choice of a specific subsampler from a list of available subsamplers. While our analysis focuses on the domain of automatic modulation classification, we believe that the detailed analysis presented in this work can apply to a wide range of domains where resource-constrained deep learning can benefit from subsampling.

\ifCLASSOPTIONcaptionsoff
  \newpage
\fi
\bibliographystyle{IEEEtran} 

\bibliography{adversarial_subsampling}

\end{document}